\documentclass[11pt,a4paper]{article}

\usepackage[round,colon]{natbib}
\providecommand{\keywords}[1]{
	\small
	\textbf{Keywords:} #1
}
\usepackage[english]{babel}
\usepackage{graphicx}%
\usepackage{multirow}%
\usepackage{amsmath,amssymb,amsfonts}%
\usepackage{amsthm}%
\usepackage{mathrsfs}%
\usepackage[title]{appendix}%
\usepackage{xcolor}%
\usepackage{textcomp}%
\usepackage{manyfoot}%
\usepackage{booktabs}%
\usepackage{algorithm}%
\usepackage{algorithmicx}%
\usepackage{algpseudocode}%
\usepackage{listings}%
\usepackage{hyperref}
\usepackage[vmargin=2cm,hmargin=2cm]{geometry}
\theoremstyle{thmstyleone}%
\theoremstyle{thmstyletwo}%

\theoremstyle{thmstylethree}%

\definecolor{blue_extractor}{RGB}{218, 232, 252}
\definecolor{purple_transformer}{RGB}{225, 213, 231}
\definecolor{orange_memory}{RGB}{255, 230, 204}
\definecolor{green_anomaly}{RGB}{213, 232, 212}
\title{Unsupervised Memory-Enhanced Video Transformers: Obstacle Detection for Autonomous Agricultural Rover}
\author{Théo Biardeau$^{1,2,*}$, Anne-Sophie Capelle-Laiz\'e$^1$, Salwan Alwan$^2$, David Helbert$^1$}
\date{$^1$ Université de Poitiers, CNRS, XLIM, France\\ $^2$ Cyclair, Pressac, France\\$^*$Corresponding author. Email: theo.biardeau@univ-poitiers.fr}
	
\begin{document}
	%%Graphical abstract
	% \begin{graphicalabstract}
		% %\includegraphics{grabs}
		% \end{graphicalabstract}
	
	%%Research highlights
	% \begin{highlights}
		% \item Label free video transformer anomaly detection method for agricultural environments.
		
		% \item Uses memory-augmented transformers for robust spatio-temporal modeling.
		
		% \item Real-time inference at 91 ms per frame on agricultural rovers.
		
		% \item Outperforms existing unsupervised methods in agricultural environments.
		
		% \end{highlights}
	
	% %% Keywords
	% \begin{keyword}
		% Anomaly Detection, Transformers, Agriculture, Autonomous rover, Security.
		
		% \end{keyword}
	\maketitle
	\begin{abstract}
	
		While autonomous rovers have become indispensable to precision farming, achieving consistent operational safety remains a critical challenge. Conventional safety sensors, such as LiDAR, fail to detect obstacles positioned below the plant canopy, posing a significant risk. While camera-based supervised learning methods can detect common objects, they perform poorly when faced with obstacles that were not present in their training data. Current unsupervised anomaly detection offers a solution by learning the normal visual patterns of an environment, but often fails for the dynamic scenes captured by a moving rover.
		
		This paper introduces Video Memory Transformers for Anomaly Detection  (VMTAD), a fully unsupervised method designed for real-time obstacle detection in dynamic agricultural scenes. VMTAD utilizes a transformer-driven architecture augmented with a dedicated memory module. This memory module leverages temporal context by processing encoded representations of preceding frames. This approach enables the system to effectively address the dynamic context caused by the robot's movement. The model is trained using only images that represent normal operation, requiring no data labels.
		
		VMTAD was rigorously evaluated on the 'Grillon' agricultural rover. On a challenging rapeseed dataset, VMTAD achieved competitive performance, reaching a 0.973 detection and 0.997 segmentation Area Under the Receiver Operating Characteristic curve. A lightweight variant provides an optimal balance of high accuracy and real-time inference (14 ms), which is critical for safety, as confirmed by our analysis of the rover's total stopping distance. Code available in the following repository: \url{https://github.com/TheoBiardeau/VMTAD}.
		
\end{abstract}	
	
	% --- KEYPOINTS / HIGHLIGHTS ---
	\section*{Keypoints}
	\begin{itemize}
		\item Proposes VMTAD, a novel fully unsupervised transformer-driven framework for real-time obstacle detection in dynamic agricultural scenes.
		\item Integrates a FIFO-based memory module and cross-attention to maintain temporal context during rover motion without quadratic complexity.
		\item Achieves competitive performance on rapeseed datasets with a detection AUROC of 0.973 and near-perfect segmentation.
		\item Features a lightweight variant (VMTAD-B0) with 14 ms inference time, ensuring safety within the robot's total stopping distance.
	\end{itemize}
	
	% --- IMPACT STATEMENT ---
	\section*{Impact}
	This research introduces a novel camera-based safety layer that overcomes the limitations of traditional LiDAR by detecting obstacles hidden beneath the plant canopy specifically during rover motion. This dedicated unsupervised method was validated through rigorous field experiments on the industrial ‘Grillon’ rover, achieving competitive performance for reliable and safe autonomous agricultural operations.

%	
%	\section*{Declarations}
%	\subsection*{Conflict of interest}
%	We declare that the authors have no Conflict of interest as defined by Springer, or other interests that might be perceived to influence the results and/or discussion reported in this paper.
%	
%	\subsection*{CRediT}
%	\textbf{Théo Biardeau}: Conceptualization, Data curation, Formal analysis, Investigation, Methodology, Validation, Visualization, Software, Writing – original draft \textbf{Anne-Sophie Capelle-Laizé}:  Methodology, Validation, Supervision, Formal analysis, Writing – review and editing. \textbf{Salwan Alwan}: Data curation, Methodology, Supervision, Formal analysis, Validation, Writing – review and editing. \textbf{David Helbert}: Methodology, Supervision, Formal analysis, Validation, Writing – review and editing.
	
	\section*{Data availability}
	Data for qualitative evaluation is available on the github: \url{https://github.com/TheoBiardeau/VMTAD}.
	Data for quantitative evaluation will be made available on request.

	\section{Introduction}

	In recent years, autonomous rovers \cite{8260901,VISENTIN2023108270,sori2018effect,UTSTUMO201836,louargant2022bipbip,QUAN202213} have become a key technology in the evolution of precision agriculture. They reduce the need for human intervention in labour-intensive tasks, such as weeding and sowing. However, ensuring the safe operation of these robots in agricultural environments where conventional safety sensors cannot be used remains a significant challenge. To fully understand the complexity of ensuring safe operation, it is important to first clarify what constitutes an obstacle in agricultural environments.
	
	Obstacles are defined as any object or condition that may obstruct the rover’s path or compromise its safe operation. Such obstacles include people, animals, machinery, agricultural waste, or fallen tree branches. These obstacles vary in size, diversity, and may be located beneath plant canopies, which poses significant challenges for detection using conventional sensors. Therefore, precise obstacle detection is essential for reducing crop damage and avoiding rover collisions or malfunctions, thereby improving overall operational efficiency.
	
	Currently, robotic systems rely on certified sensors, such as LiDAR and radar, to detect obstacles and stop operations when necessary. However, these solutions have a significant drawback: sensors must be mounted above the crop to avoid false positives, which prevents them from detecting obstacles below the plant canopy.

	\begin{figure}[!h]
		\centering
		\includegraphics[width=1.\linewidth]{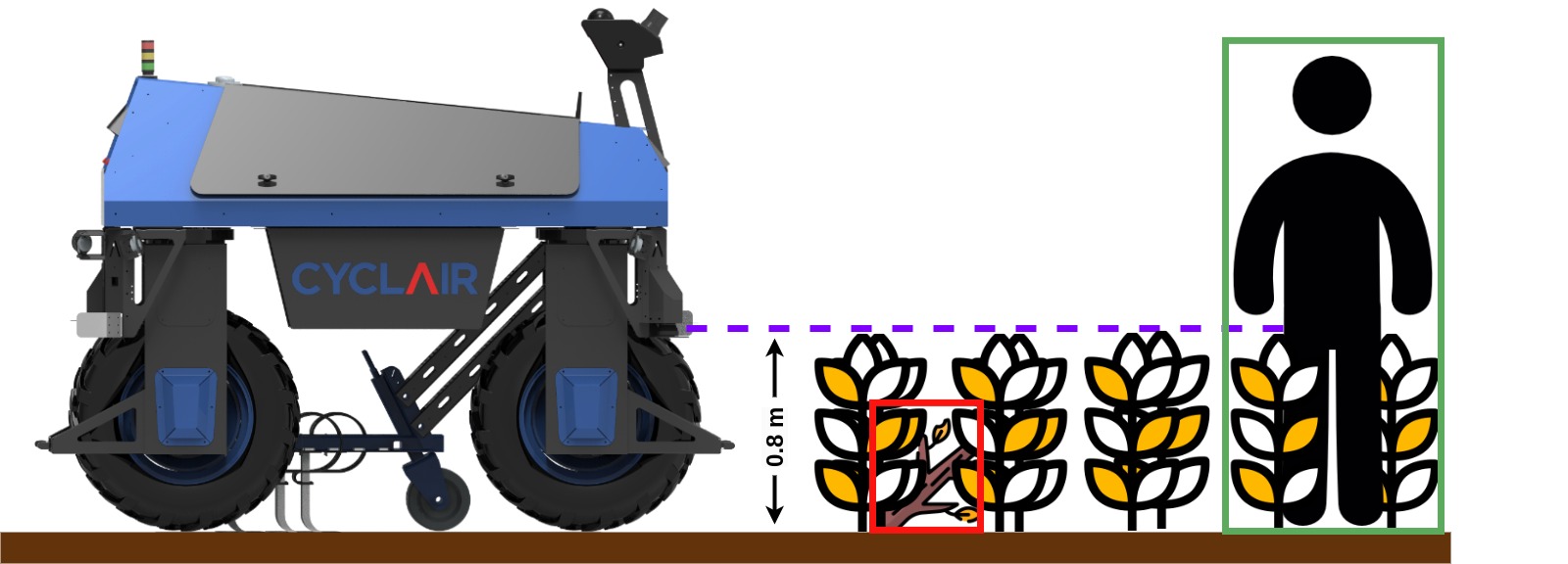}
		\label{fig: Grillon}
		\caption{Illustration of the obstacle detection challenge for an agricultural rover. Large obstacles indicated by the green box can be easily detected by safety LiDAR or radar, with the minimum detection height indicated by the purple line. However, smaller obstacles such as those by the red box cannot be detected by these sensors because they are smaller than plants. Only a method using more complex sensors, such as an RGB camera, can detect these obstacles.}
	\end{figure}
	
	Consider the 'Grillon' rover platform developed by Cyclair (Pressac, France) Figure \ref{fig: Grillon} operates autonomously at 1.2 $m.s^{-1}$  with a working width of 3 meters and relies on a LiDAR sensor mounted at 80 cm to detect obstacles. The weeding task is the reason for this configuration. However, objects measuring less than 80 cm will not be detected.

	A promising alternative is the integration of RGB cameras, which provide rich and high-resolution visual information. One potential approach involves constructing an obstacle dataset and employing supervised learning techniques with models such as YOLO \cite{Redmon_2016_CVPR}. Although this strategy can reliably detect common obstacles such as people or tree branches, it tends to perform poorly when faced with anomalies that were not represented in the training data. This limitation highlights the need for methods that can generalise to unseen scenarios.
	
	By redefining obstacles as deviations from the "normal" rather than predefined categories, unsupervised anomaly detection addresses this challenge. Instead of relying on exhaustive labeled datasets, it learns the normal visual patterns of the environment and identifies deviations from this baseline as potential obstacles. This enables the system to recognize previously unseen objects or conditions without prior examples, making it particularly well-suited for the unpredictable and dynamic nature of agricultural fields. However, the methods currently used in agriculture \cite{s23167285} and more generally in the video surveillance industry do not take into account the movement of the acquisition system, which is our case since the rover is moving. 
	
	To overcome these limitations, this paper introduces VMTAD: a fully unsupervised, transformer-driven framework designed for dynamic agricultural scenes with real-time inference capabilities. VMTAD exploits the spatio-temporal modelling capabilities of Transformer architectures \cite{NIPS2017_3f5ee243} augmented with a dedicated temporal memory module, enabling detection under moving cameras. VMTAD does not require any labels; only images representing normality are necessary to train the model.
	
	This paper is an extended version of \cite{biardeau2025unsupervised}, in which the original VMTAD architecture was introduced. In this paper, we make several contributions, which are as follows:
	
	\begin{itemize}
		\item Detailed methodology description.
		\item Two model versions, VMTAD-B5 and VMTAD-B0, are introduced and compared, with EfficientNetB5 and EfficientNetB0 being utilised as their respective feature extractors.
		\item A supplementary study is added to evaluate the impact of specific components, including the influence of the memory FIFO queue size and the memory decoder length.
		\item A rigorous assessment of the total stopping distance, integrating systemic latency with mechanical braking dynamics.
		\item Introduction of new results, notably more granular performance results, with detection  and segmentation metrics explicitly separated, and updated Area Under the Receiver Operating Characteristic curve (AUROC) scores reported on the Corn and Rapeseed datasets.
		
	\end{itemize}
	
	To contextualize the development of VMTAD, it is essential to examine the evolution of anomaly detection (AD) methods, particularly the shift from supervised object detection to unsupervised frameworks capable of handling the unpredictability of field operations.
	
	Although AD is essential for ensuring the safety of robots, traditional deep learning approaches often require exhaustive manual labelling. To bypass this requirement, research has shifted towards self-supervision. For instance, a self-supervised method  \cite{8963641} for robotic wheelchairs has been developed that uses RGB-D data to automatically generate labels for drivable areas and road anomalies. While effective for structured environments, this approach is less suitable for agriculture, as the depth sensors used often fail in direct sunlight, and the system is designed to detect ground-level anomalies like curbs, not complex obstacles such as farm equipment or obstacle partially hidden by crops.
	For dynamic indoor navigation, \cite{10930535} proposed a framework using an unsupervised autoencoder to learn and reconstruct normal scenes, identifying anomalies through high reconstruction errors. Critically, they integrated conformal prediction to provide statistically guaranteed uncertainty measures for each detection. The reliance of this method on indoor training data renders it ineffective for the unpredictable conditions of an agricultural landscape. The visual diversity and constant changes in an agricultural field make it difficult to establish a stable "normal" scene, which is the core assumption of the model.
	In challenging outdoor settings where standard classifiers may fail on out-of-distribution data, \cite{8786197} introduced a self-supervised system that learns normal terrain from the robot's experience using multi-modal data. Their use of a normalizing flow model \cite{kingma_glow_2018} allows the system to robustly identify unexpected hazards. This approach focuses on the traversability of the terrain itself (e.g., soil vs. mud) rather than detecting discrete obstacles located on top of it. Because it learns retrospectively from areas the robot has already passed over, it cannot proactively identify obstacles before interaction.
	Similarly,  \cite{mattamala2025wild} presented an online self-supervised system for visual traversability using features from pre-trained models that adapt from proprioceptive feedback. The primary limitation of this method for agricultural safety is its reliance on physical interaction to generate supervision signals. The robot must attempt to traverse an area to learn if it is an obstacle, making it fundamentally unsafe for detecting objects like people, animals, or machinery.
	
	To the best of our knowledge, the only work addressing this issue in the agricultural context is \cite{s23167285}, which applies AD to an agricultural environment. Their approach combines a supervised YOLO-based detector for common obstacles with a semi-supervised convolutional autoencoder for rare anomalies. Although effective, this method depends on labeled data and strong prior assumptions about obstacle types, resulting in high annotation costs and potential risks when certain obstacle classes are missing from the training set. Moreover, these approaches do not leverage recent advances in industrial AD, where significant research efforts have been devoted to improving robustness and generalization. This reliance on prior assumptions in agriculture stands in stark contrast to the evolution of industrial AD, where the focus has shifted toward unsupervised robustness and generalizable feature representation.

	In industrial AD, PatchCore \cite{Roth_2022_CVPR} obtains impressive results by matching multi-scale patch embeddings against a compact memory bank of normal patterns in the MVTec \cite{bergmann_mvtec_2021} benchmark. Subsequent self-supervised method, such as SimpleNet \cite{Liu_2023_CVPR} retain competitive area under the ROC curve AUROC and inference time. Despite their speed and accuracy, these methods represent one major direction in AD research focused on local feature. Another significant research trend has centered on Transformer-based architectures, which are designed to model the global context of an image \cite{AMMAR2026103517}.
	
	Recently, transformers architectures have demonstrated superior performance over convolutional autoencoders (CAE) in AD and segmentation tasks. Notable examples include AnoViT \cite{9765986} and VT‑ADL \cite{9576231} replace the encoder with a pre‑trained ViT \cite{dosovitskiy2020image} and show consistent gains over CAE baselines. ADTR \cite{ADTR_you} generalises this idea into a full Transformer encoder-decoder, achieving state‑of‑the‑art accuracy with a pure deep‑learning pipeline. GeneralAD \cite{GeneralAD_cite} further demonstrates that self-supervised transformer AD models have competitive performance that SimpleNet \cite{Liu_2023_CVPR} but at the cost of extensive inference time. Despite their accuracy, all of them operate on unique, fixed frames, implicitly assuming fixed viewpoints and negligible movement assumptions, which causes difficulties on mobile platforms such as agricultural rovers. It is therefore necessary to explore methods that include a temporal aspect like video transformers.
	
	Applying Transformers to video data remains a significant challenge due to the quadratic growth of self-attention complexity relative to the number of tokens \cite{10041724}. To mitigate this, current state-of-the-art (SOTA) methods frequently rely on computational compromises that are often incompatible with the constraints of mobile robotics in unstructured environments. For instance, tubelet-based approaches \cite{jin2022anomaly} reduce overhead by grouping neighboring pixels, yet they depend on overly simplistic temporal assumptions. Similarly, Transanomaly \cite{TransAnomaly} utilizes a CNN-based encoder-decoder for primary spatial reconstruction; however, its subsequent reliance on sequential spatial and temporal transformers operating strictly within the latent space results in a loss of fine-grained detail. This makes it particularly difficult to identify the subtle or small-scale anomalies frequently encountered in agricultural contexts. Following a similar logic, \cite{Tran2024Traffic} proposed a multi-step unsupervised framework designed for aerial surveillance. Their architecture employs a spatial transformer followed by a temporal transformer to capture long-range object relationships. The model primarily identifies anomalies based on high future frame reconstruction errors. While such a multi-step approach is effective for common traffic scenarios, such as highways or roundabouts, it can still struggle with the complex rotation angles of the camera and the agricultural environment.
	
	In contrast, we propose VMTAD, a dedicated architecture designed for real-time AD on moving platforms. While we leverage established components like transformers and memory modules, our scientific contribution lies in a novel architectural synergy tailored for motion compensation and high-fidelity reconstruction dedicated at agricultural AD. Our method distinguishes itself in three fundamental ways:
	
	\begin{itemize}
		\item Explicit Temporal Ordering: Unlike learnable latent memories that can lose chronological consistency, we employ a FIFO (First-In, First-Out) queue of encoded frames. This explicitly preserves the temporal sequence, a critical requirement for compensating for the motion of agricultural rovers.
		\item Linear-Scaling Cross-Attention: We bypass the quadratic cost of standard attention by using cross-attention between the current frame and the FIFO memory. This achieves linear scaling with memory length, enabling the use of significantly larger temporal windows while maintaining the real-time inference required for autonomous navigation.
		\item Pure-Transformer Reconstruction: Departing from hybrid CNN-Transformer models, VMTAD maintains full frame-level spatial resolution through a dual spatial-temporal fusion directly within the Transformer layers. This avoids the information loss inherent in latent compression, ensuring the detection of small-scale agricultural hazards.
	\end{itemize}
	
	\section{Method}  \label{Method}
	
\begin{figure}[!ht]
		
		\includegraphics[trim={1.2cm 21cm 0.5cm 1cm},clip,width=1.\textwidth]{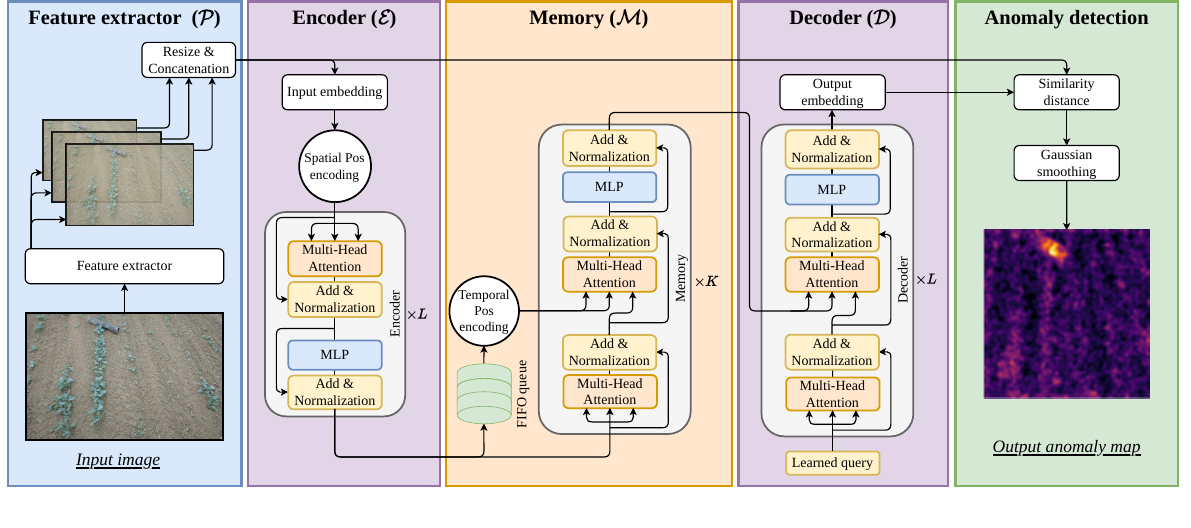}
		\caption{
			Overview of the proposed VMTAD. The pretrained feature extractor \(\mathcal{P}\) generates hierarchical features from RGB image, which are resized and concatenated into a unified multi-scale feature representation. These feature maps are then embedded and input into the transformer-based encoder \(\mathcal{E}\) that employs multi-head attention to model spatial context. Memory module  \(\mathcal{M}\) uses cross-attention mechanisms between the current encoded frame and historical frames contained in FIFO queue.  The decoder \(\mathcal{D}\) reconstructs the input from learned query and the memory module output. The anomaly map is generated by anomaly detection block that comparing the model decoder output and  encoder input via cosine similarity.}
		\label{fig:VMTAD}
		\end{figure}

	This section provides a detailed description of our proposed method, VMTAD, as illustrated in Figure~\ref{fig:VMTAD}. It consists of 5 main components: Feature Extractor \(\mathcal{P}\), a transformer-based encoder \(\mathcal{E}\), a transformer-based memory \(\mathcal{M}\),  a transformer-based decoder \(\mathcal{D}\), and an anomaly detection module. These modules are described in detail below.
	
	\subsection{Problem Formulation}
	The aim of video AD is to identify regions of a video frame that deviate from learnt normal patterns. Given a sequence of \(I\) frames \(\{I_1, \dots, I_t\}\), the task is to detect anomalies in \(I_t\) by comparing its spatial and temporal characteristics with normal patterns acquired during training. VMTAD uses a reconstruction-based approach to model normality: it is trained to reconstruct normal data, and during inference, reconstruction errors on normal data are low, whereas reconstruction errors on abnormal data (e.g. obstacle presence) are high.
	
	\subsection{Feature Extractor}
	The feature extractor $\mathcal{P}$, pretrained on large and augmented datasets like ImageNet1k \cite{5206848}, produces rich and versatile representations that are especially beneficial for AD. By leveraging frozen pretrained extractors, AD methods can more accurately model the distribution of normal data while, maintaining robustness against domain shifts, thereby enhancing overall detection performance.
	
	EfficientNet \cite{pmlr-v97-tan19a}  employed as the backbone to generate feature maps. EfficientNet is a fast feature extractor capable of producing  compact and rich feature maps, which makes it a perfect candidate for an agricultural application with transformer model. Given an input image \( I \in \mathbb{R}^{H \times W \times C} \), where $H$, $W$, and $C$ denote the height, width, and number of channels of the image respectively, \(\mathcal{P}\) extracts hierarchical feature maps from its internal layers. The extracted features are resized to the same resolution and concatenated into a multi-scale feature map \(\mathcal{P}(I) = F \in \mathbb{R}^{H_1 \times W_1 \times C_1} \), where $H_1$, $W_1$, and $C_1$ denote the height, width, and number of channels of the features map. This multi-scale feature map serves as input to the subsequent encoder, allowing it to exploit both fine-grained and high-level  information.

	\subsection{Transformers Encoder-Decoder}
	Our model architecture is based on the paper introducing transformers \cite{NIPS2017_3f5ee243} and adapted for feature processing. The input feature map \(F\) is first flattened into tokens to form matrix \(\mathbf{F} \in \mathbb{R}^{H_1W_1 \times C_1} \), where each token \(\mathbf{f} \in \mathbb{R}^{C1} \) corresponds to a spatial location. These tokens are projected using a trainable weight matrix. To preserve spatial position information, sine and cosine functions are used, as in the transformers' original paper \cite{NIPS2017_3f5ee243}.
	
	The transformer encoder uses a multi-head attention (MHA) mechanism to capture contextual relationships by computing self-attention on \(\mathbf{f}\). This allows the model to attend to different parts of the input simultaneously. The MHA output is normalized and combined with a skip connection from the input before being processed by a multi-layer perceptron (MLP). This sequence of operations is repeated across layers, with each layer receiving input from the previous layer except the first, which receives input from the embedding step. After passing all tokens through $L$ layers of encoders, we get \(\mathbf{F}_{encoded}\). Considering only the self-attention and MLP components, the complexity of the encoder can be expressed by the following equation:
	
	\begin{equation}
		T(\mathcal{E}(\mathbf{F})) = O(L \cdot ((H_1W_1)^{2} \cdot C_1 + H_1W_1 \cdot C_1^{2})).
	\end{equation}
	
	In other words, the complexity is quadratic with respect to the number of features and the depth of each feature, and linear with respect to the number of layers. Therefore, reducing both the number of features and the information they contain is important for achieving time efficiency. By selecting EfficientNet as the feature extractor, the model produces compact yet informative feature maps that fulfil this requirement. This architectural synergy ensures that the transformer's computational burden remains low, enabling the necessary inference speed to maintain a safe stopping distance during rover motion.

	The transformer decoder generates output tokens \(\hat{\mathbf{f}}\) by considering both previous outputs and the encoded memory. It starts with learnable queries that are processed by self-attention. In subsequent layers, these queries are replaced by the output of the previous layer. The self-attention output is normalized and combined with a skip link. Cross-attention is then applied, combining the memory output with the self-attention result to ensure connections across sequences. Layer normalization, skip connections, and a MLP refine the output through multiple L decoder layers, iteratively improving the representation across layers. The final tokens of the decoder are projected into their original space and reshaped to form \(\hat{F} \in \mathbb{R}^{H_1 \times W_1 \times C_1} \). The complexity of decoders, considering only self-attention and cross-attention by these importances, we obtain the following equation:
	\begin{equation}
		T(\mathcal{D}(\mathbf{F}_{encoded})) = O(L \cdot (2 \cdot (H_1W_1)^{2} \cdot C_1 + H_1W_1 \cdot C_1^{2})).
	\end{equation}
	
	Like encoders, decoders have quadratic complexity related to the number of features and the information contained in each feature, and linear complexity related to the number of encoder layers.

	\subsection{Memory}
	The memory module is designed to capture relevant temporal information from previous frames. This allows the model to reason not only about what is seen in the current frame, but also about what has been observed recently. By storing previous encoding feature maps, the memory acts as a short-term context buffer, helping the model to detect subtle anomalies. This temporal continuity is especially important in agricultural environments where the robot is in motion, introducing continuous changes in the scene.
	
	The memory module employs a Transformer decoder architecture consisting of $K$ layers to effectively model the relationships between the current encoded frame and previous frames stored in a FIFO queue. This FIFO structure is advantageous as it preserves the strict temporal ordering of the sequence while maintaining the encoded frames at full spatial resolution. In the initial layer, the most recent encoded frame is processed via self-attention. In subsequent layers, this input is replaced by the refined output of the preceding layer. Each layer follows a systematic refinement process: the self-attention output is normalized and integrated through a skip connection, followed by a cross-attention mechanism that computes dependencies between the tokens and features stored in the FIFO queue. Finally, the representation is further processed by a MLP, complemented by layer normalization and residual connections. To maintain temporal context, we integrate sine and cosine temporal positional encodings, adapted specifically for the memory architecture. The computational cost of this module is dictated by the transformer decoder's structure, leading to an algorithmic complexity defined as follows:
	\begin{multline}
		\label{eq: complexity memory}
		T(\mathcal{M}(\mathbf{F}_{FIFO},\mathbf{F}_{encoded} )) = O(K \cdot ((H_1W_1)^{2} \cdot C_1 + \\ N \cdot (H_1W_1)^{2} \cdot C_1 +  H_1W_1 \cdot C_1^{2}))),
	\end{multline}
	where \(N\) denotes the number of elements in FIFO queue. Like decoder, the memory's algorithmic complexity is quadratic relative to the spatial resolution $(H_1W_1)$ and linear relative to the number of decoder layers $K$. Crucially, the cross-attention mechanism ensures that the complexity scales linearly with the number of encoded frames in memory ($N$). This property allows the system to leverage a larger temporal context without the prohibitive computational cost. In the context of AD in agricultural robotics, this provides the robot with a significant time window to process the concept of movement and thereby improve performance.
	
	\subsection{Objective Function}
	We select cosine similarity as the reconstruction objective for two main reasons. First, it is well-suited for high-dimensional feature spaces, where angular relationships are often more informative than vector magnitudes. This property is widely exploited in SOTA architectures such as CLIP \cite{radford2021learning}, where cosine similarity enables robust alignment of semantically meaningful embeddings. In our case, and in line with the reconstruction paradigm of VMTAD, we leverage this property to enforce consistency between input feature maps and their reconstructions in the latent space.
	
	Second, cosine similarity is invariant to feature scaling, unlike $L_1$ or $L_2$ losses which are sensitive to vector norms. This makes it more robust to illumination variations, a key factor in agricultural environments where lighting conditions can change rapidly. As a result, the model is encouraged to detect anomalies as semantic deviations rather than low-level intensity shifts.
	
	The cosine similarity between input features $F$ and reconstructed features $\hat{F}$ is defined as:
		\begin{equation}
			\label{eq: cos}
		\mathcal{L} = 1 - \frac{F \cdot \hat{F}}{\|F\| \|\hat{F}\|}.
	\end{equation}

	\subsection{Anomaly Detection}
	Anomaly score maps, denoted as \(M_{\text{raw}}\), are computed using the cosine similarity Equation \ref{eq: cos} between each vector of the feature maps. To reduce noise and minimize false positives, \(M_{\text{raw}}\) is convolved with a 2D Gaussian kernel \(G\), resulting in the refined map \(M\):
	
	\begin{equation}
		M = M_{\text{raw}} \ast G,
	\end{equation} where \(\ast\) represents the convolution operation. 
	
	In anomaly segmentation, the full map $M$ is used to localize anomalies across the image. In contrast, AD relies only on the maximum score $d_{\text{detect}}$ extracted from $M$:
	
	\begin{equation}
		d_{\text{detect}} = \max(M).
	\end{equation}

	\section{Material and experimentation}
	
	\subsection{Robotic Platform}
	
	\begin{figure}[ht]
		\centering
		\includegraphics[width=1\linewidth]{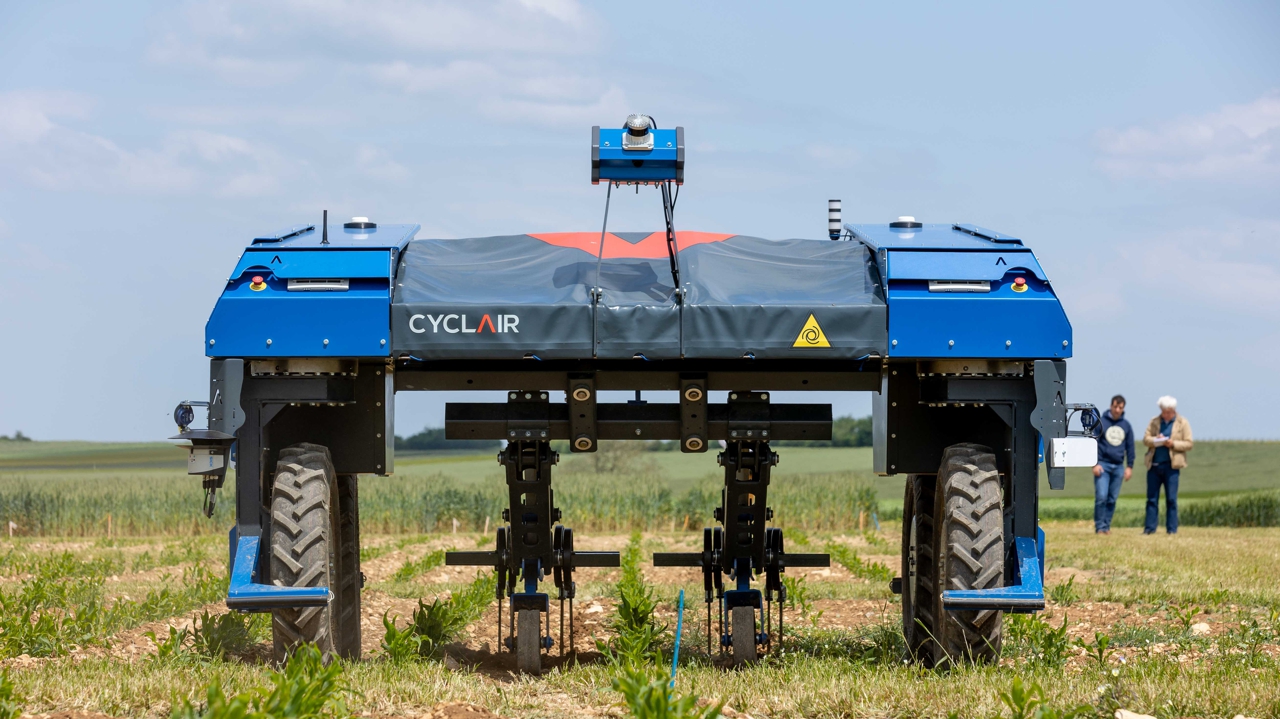}
		\caption{The weeding rover platform, a 2 ton, 3 meter autonomous agricultural rover designed specifically for weeding operations. It uses artificial intelligence and GNSS to navigate crop fields.}
		\label{fig: Grillon rover}
	\end{figure}

	To ensure evaluation of the method under realistic agricultural conditions, the 'Grillon' rover platform was utilized, as shown in Figure. \ref{fig: Grillon rover}. This rover is specifically designed for weed control tasks, weighs approximately two tons, and has an operational width of 3 meters. It is equipped with a high dynamic range (HDR) camera tailored for navigation purposes which captures detailed visual data crucial for our AD task. During data acquisition, the rover executed systematic back-and-forth movements at speeds ranging from 0 to 1.2  $m.s^{-1}$. Inference is performed on an Nvidia RTX A2000 laptop.
	
	\subsection{Dataset}
	
	The dataset used for our experiments was exclusively collected in rapeseed fields in Pressac, France. Supplementary evaluations were conducted in corn fields in Pressac, France. These crops represent the agricultural weeding needs of farmers in the region.  During data acquisition, the Grillon rover executed systematic back-and-forth movements at speeds ranging from 0 to 1.2 \(m.s^{-1}\). The training dataset comprised 9 sequences, each containing 600 obstacle-free images captured in rapeseed fields. For evaluation, the dataset included 5 sequences of 600 images from rapeseed fields and 13 sequences of 600 images from corn fields. A high variety of obstacles were present in the evaluation data, including agricultural tarpaulins, plastic debris, wooden pieces, fence posts, irrigation hoses, and occasional human presence, across both environments. The evaluation dataset was meticulously annotated using the SAM2 \cite{ravi2024sam2} annotation tool to ensure precise labelling of obstacles.
	
	A detailed demonstration video is available in the code repository, showcasing the real-world deployment, including AD results. Although the complete video dataset is confidential, a representative sequence has been released along with the full VMTAD code to enable quantitative validation of the results.
	
	\subsection{Model Setup}
	
	In this subsection, the different operational modes of the model are described:
	
	\textbf{Train Mode:} Processes \(I+1\) images where the first \(I\) images are pre-processed, encoded, and cached in FIFO queue to avoid redundant calculations, while the \((I+1)\)th image is fully processed to update the model.
	
	\textbf{Stream Mode:} For inference, the first frame initializes memory by duplicating its encoded features \(N\) times. Each subsequent frame is processed individually, with its encoded features replacing the oldest in a FIFO queue.
	
	For feature extraction, VMTAD-B5 and VMTAD-B0 employ EfficientNet-B5 and EfficientNet-B0 architectures \cite{pmlr-v97-tan19a} pretrained on ImageNet \cite{5206848}. By concatenating features from the first three layers, the models aggregate hierarchical information ranging from low-level spatial textures to high-level semantic patterns, yielding output dimensions of $96\times 96\times 128$ for VMTAD-B5 and $56\times 56\times 80$ for VMTAD-B0. Training is conducted over 50 epochs with a learning rate of $1 \times 10^{-4}$ and a batch size of 2. The Transformer hyperparameters include a latent dimension of $C_1=128$ (resp. $80$), encoder-decoder depth of $L=4$, MLP dimension of 1024, and 8 attention heads. These hyperparameters are conventionally used in standard transformers and transformer-based AD \cite{dosovitskiy2020image,ADTR_you}; they represent a heuristic trade-off between reconstruction capacity, overfitting risk, and computational efficiency, a critical balance for our application. The memory depth is set to $K=2$ with a FIFO memory length of $N=2$. Given that this mechanism represents one of our major contributions, the specific influence of FIFO size and memory length is further analyzed in the supplementary study. Following the methodology of \cite{GeneralAD_cite}, the anomaly score is evaluated at each epoch. To mitigate overfitting, an early stopping mechanism with a patience of 10 epochs is implemented, and a Gaussian kernel of size 3 is ultimately applied to the output anomaly maps.
	
	\subsection{Concurrent methods}
	VMTAD was compared against unsupervised methods commonly used in agriculture, such as ConvAE, VQ-VAE, and MemAE. The experimental parameters established by \cite{s22103608} were employed for the evaluation of these methods. An $L_2$ norm calculation between the input and output images was utilized for AD, with the application of a Gaussian blur identical to the one used for VMTAD. Transitioning from these reconstruction-based agricultural models to more diverse architectural paradigms, we also considered top-tier industrial solutions. Regarding general SOTA AD methods, PatchCore \cite{Roth_2022_CVPR} was selected for its high performance on the MVTec-AD dataset \cite{bergmann_mvtec_2021} and its operational paradigm, which utilizes a memory bank derived from a feature coreset generated by a backbone. We limited the memory bank size to 9,216 elements to maintain a consistent inference time, while keeping the same hyperparameters proposed in the original publication. GeneralAD \cite{GeneralAD_cite} was also included for its excellent performance on MVTec-AD and its innovative use of transformers. We followed the parameters proposed in the original publication , tuning only the noise ratio for a fair comparison. We opted for the random noise configuration, as it yielded better results than other modes in our context. Finally, SimpleNet \cite{Liu_2023_CVPR} was chosen for its exceptional speed relative to its performance. Similar hyperparameters were used, omitting the default noise which was customized to ensure a fair comparison. To ensure architectural fairness, all methods (VMTAD, PatchCore, SimpleNet, GeneralAD) use the same feature extractor (EfficientNet-B5) and identical input preprocessing. This eliminates performance differences stemming from backbone capacity and focuses the comparison on the AD mechanisms.

	\section{Results}
	\label{sec:results}
	
	The evaluation of the VMTAD method was conducted on two agricultural datasets (Rapeseed and Corn) and compared against several SOTA approaches. Performance was quantified using the Area Under the Receiver Operating Characteristic curve (AUROC) for both detection (Det.) and segmentation (Seg.), alongside inference latency measurements.
	
	\subsection{Performance on Rapeseed Dataset}
	Quantitative results for the Rapeseed dataset are presented in Table~\ref{tab:results_rapseed}. The VMTAD-B5 variant achieved an AUROC score of 0.973 for detection and 0.997 for segmentation. The inference time for this model was measured at 64 ms. The lightweight version, VMTAD-B0, reached a segmentation performance of 0.996 with a detection of  0.950 with a reduced latency of 14 ms. In comparison, reconstruction-based methods (CAE, VQ-VAE, MemAE) \cite{s22103608} exhibited lower latencies (4 to 5 ms) but lower detection scores, not exceeding 0.775. Among SOTA industrial methods, PatchCore \cite{Roth_2022_CVPR} achieved the highest detection score (0.977) with a contained inference time of $36$ms, while GeneralAD \cite{GeneralAD_cite} presented the highest latency (101 ms) with the highest segmentation score (0.996) of industrial method. SimpleNet \cite{Liu_2023_CVPR} is the faster industrial model with a inference time of $17$ms but with the worst segmentation (0.976) and detection (0.938) results among industrial AD methods.

	\begin{table}[ht]
		\centering
		\caption{AUROC performance on rapeseed datasets. In \textbf{bold} the best result and \underline{underline} the second-best result.}
		\label{tab:results_rapseed}
		\begin{tabular}{lccc}
			\toprule
			Method    & Det. $\uparrow$     & Seg. $\uparrow$   & Time (ms) $\downarrow$ \\ 
			\midrule
			CAE       & 0.697               & 0.718             & \textbf{4} \\
			VQ-VAE    & 0.775               & 0.822             & \underline{5} \\
			MemAE     & 0.761               & 0.751            	& \textbf{5} \\
			\midrule
			SimpleNet & 0.938               & 0.976            	& 17 \\
			GeneralAD & 0.960               &\underline{0.996}  & 101 \\            
			PatchCore & \textbf{0.977}      & 0.987             & 36 \\
			\midrule
			\midrule
			VMTAD-B5  & \underline{0.973}   & \textbf{0.997}    & 64 \\
			VMTAD-B0  & 0.950               & \underline{0.996} & 14 \\
			\bottomrule
		\end{tabular}
	\end{table}
	
	\subsection{Generalization to Corn Dataset}
	The performance of models trained on Rapeseed and tested on Corn is reported in Table~\ref{tab:results_corn}. A decline in detection scores was observed for the VMTAD variants, with 0.880 for VMTAD-B5 and 0.823 for VMTAD-B0. However, segmentation scores remained above 0.960. Agricutlural method (CAE, VQ-VAE, MemAE) exhibit a very low detection and segmentations AUROC performance with AUROC not exceeding respectively 0.640 and 0.698.
	SOTA industrial methods such as SimpleNet and PatchCore maintained higher detection and segmentation AUROC on this second domain.
	
	\begin{table}[ht]
		\caption{AUROC performance on corn datasets (trained on rapeseed). In \textbf{bold} the best result and \underline{underline} the second-best result.}
		\label{tab:results_corn}
		\centering
		\begin{tabular}{lcc}
			\hline
			Method      & Det. $\uparrow$       & Seg. $\uparrow$\\ 
			\hline
			CAE         & 0.537                 & 0.626  \\
			VQ-VAE      & 0.640                 & 0.604 \\
			MemAE       & 0.606                 & 0.698  \\
			\hline
			SimpleNet   & \textbf{0.919}        & \underline{0.975} \\
			GeneralAD   & 0.884                 & 0.956 \\
			PatchCore   & \underline{0.917}     & \textbf{0.983} \\
			\hline
			\hline
			VMTAD-B5    & 0.880                 & 0.971  \\
			VMTAD-B0    & 0.823                 & 0.960 \\
		\end{tabular}
	\end{table}
	
	\subsection{Qualitative Analysis}
\begin{figure*}[!ht]
		\centering
		% Answer: [trim={left bottom right top},clip]
		\includegraphics[trim={5cm 10cm 5.7cm 9.5cm},clip,width=0.99\linewidth]{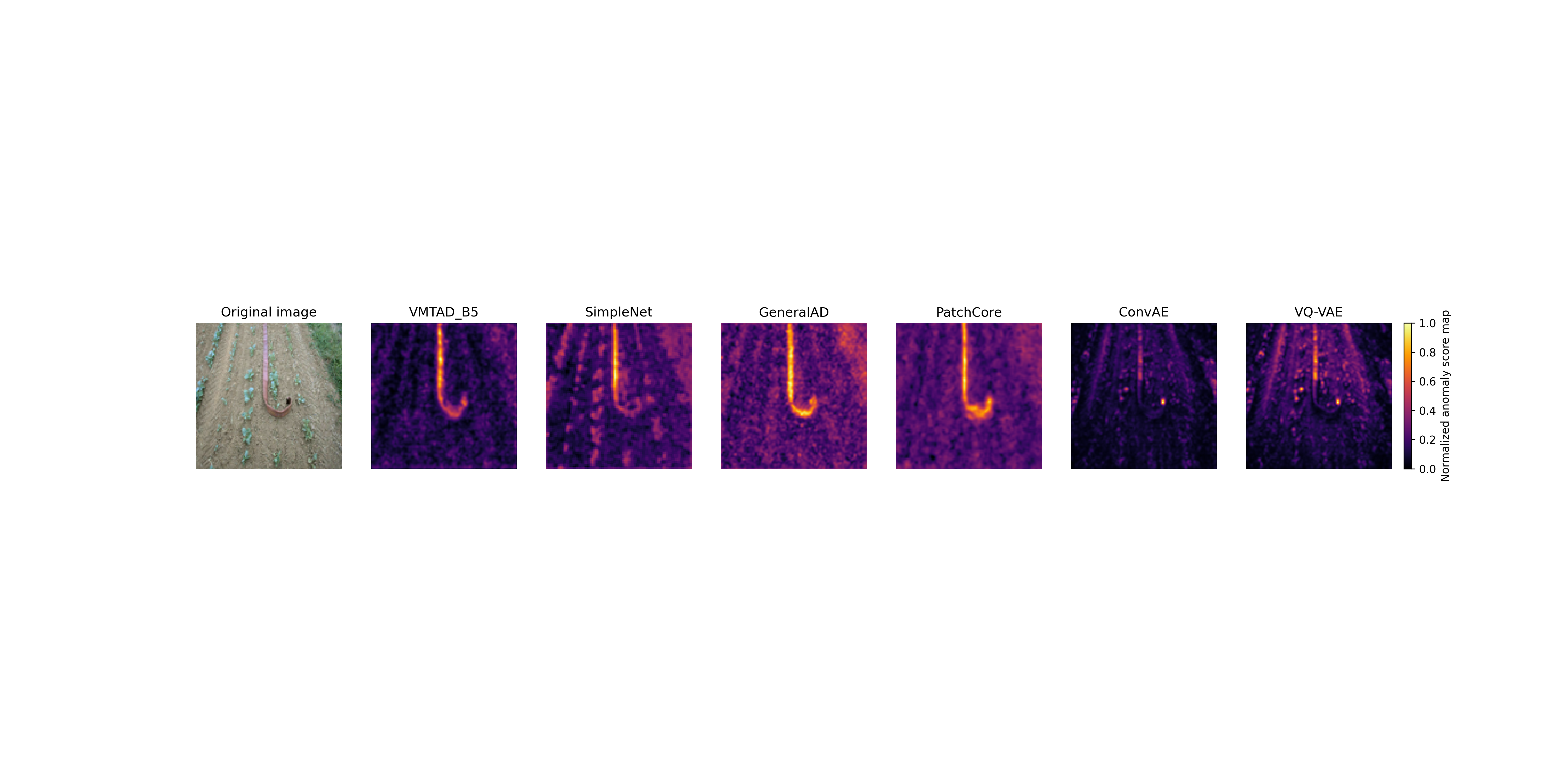}
		\includegraphics[trim={5cm 10cm 5.7cm 10.4cm},clip,width=0.99\linewidth]{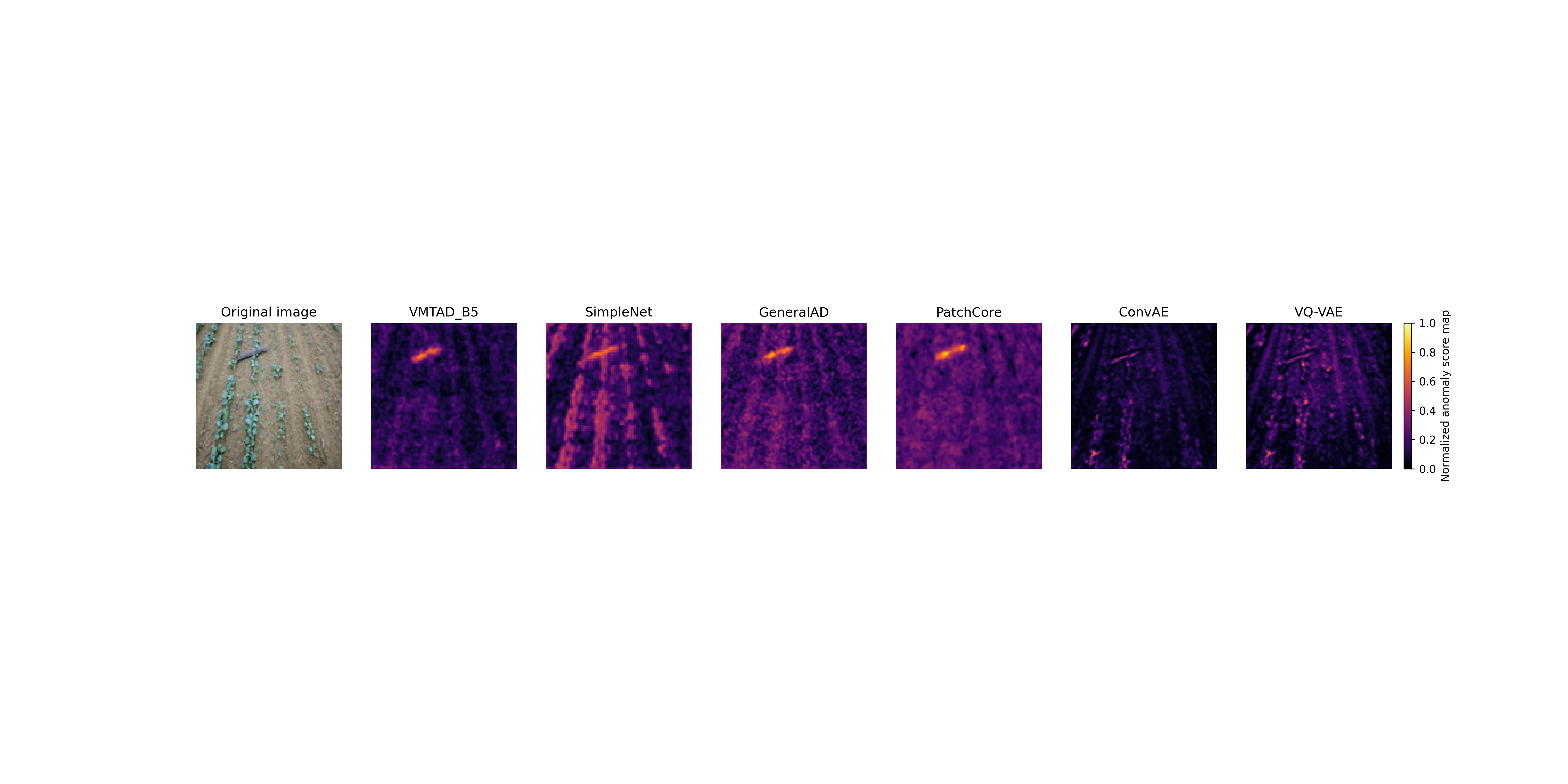}
		\includegraphics[trim={5cm 10cm 5.7cm 10.4cm},clip,width=0.99\linewidth]{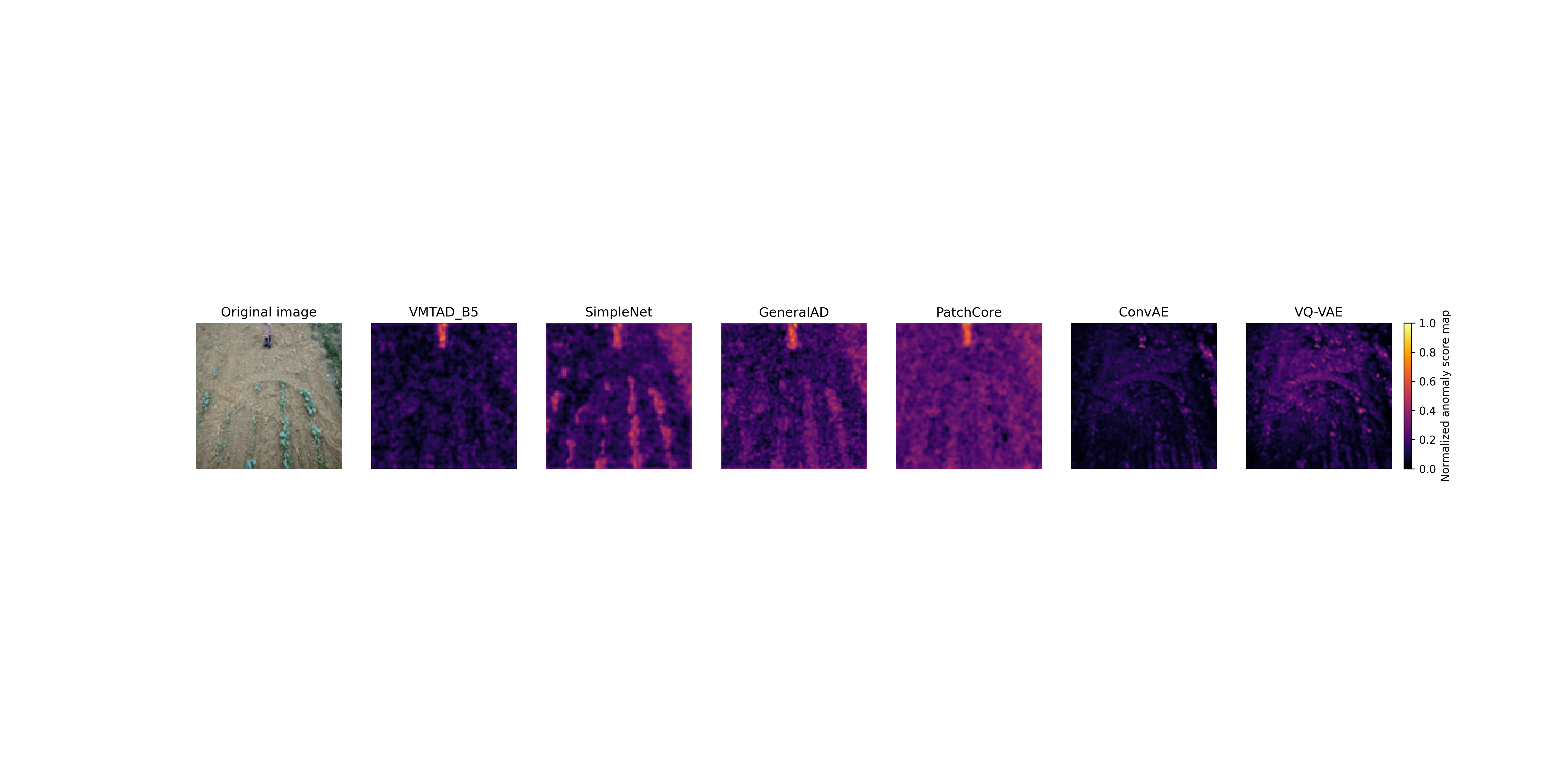}
		\includegraphics[trim={5cm 10cm 5.7cm 10.4cm},clip,width=0.99\linewidth]{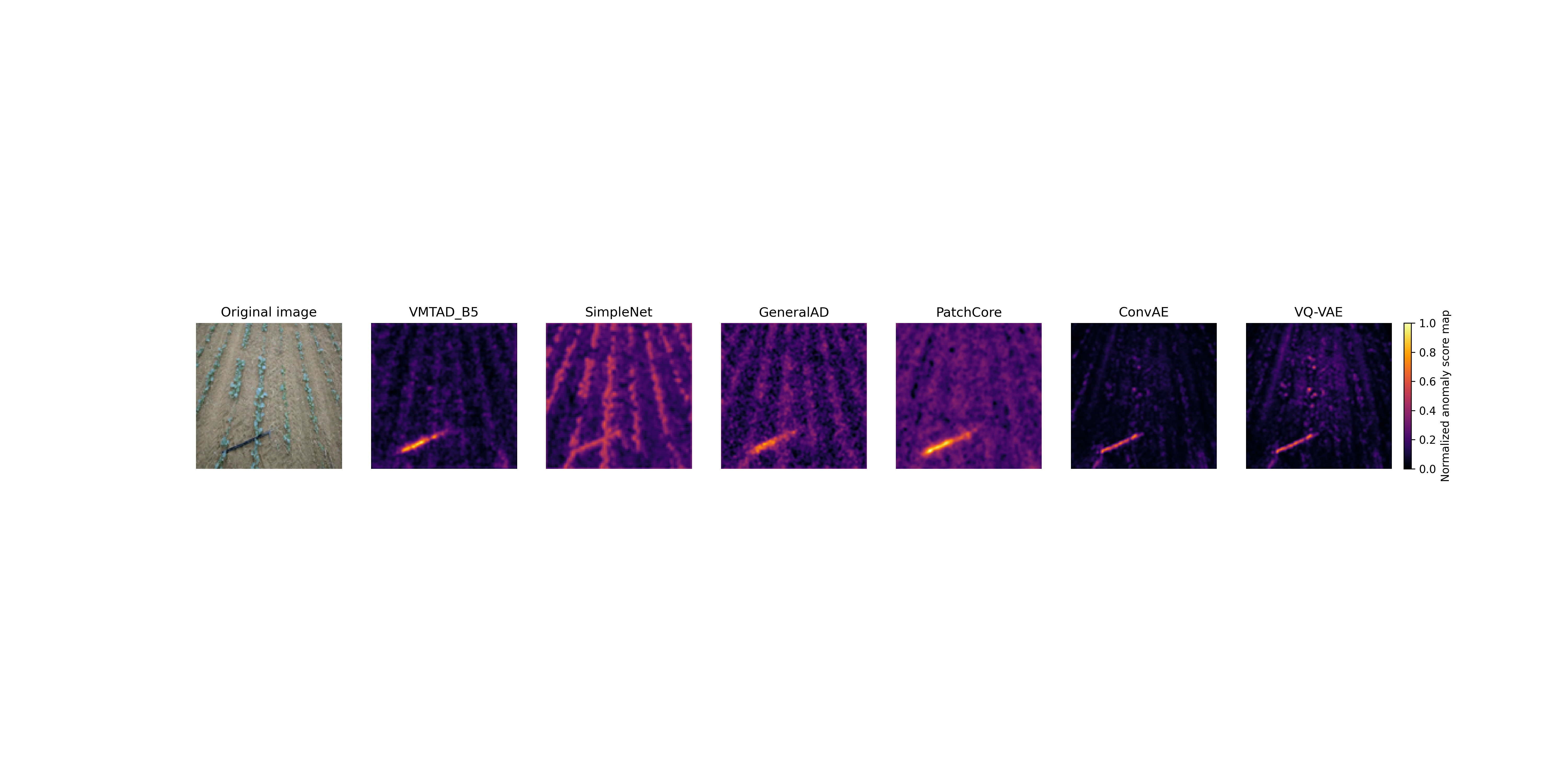}
		\includegraphics[trim={5cm 10cm 5.7cm 10.4cm},clip,width=0.99\linewidth]{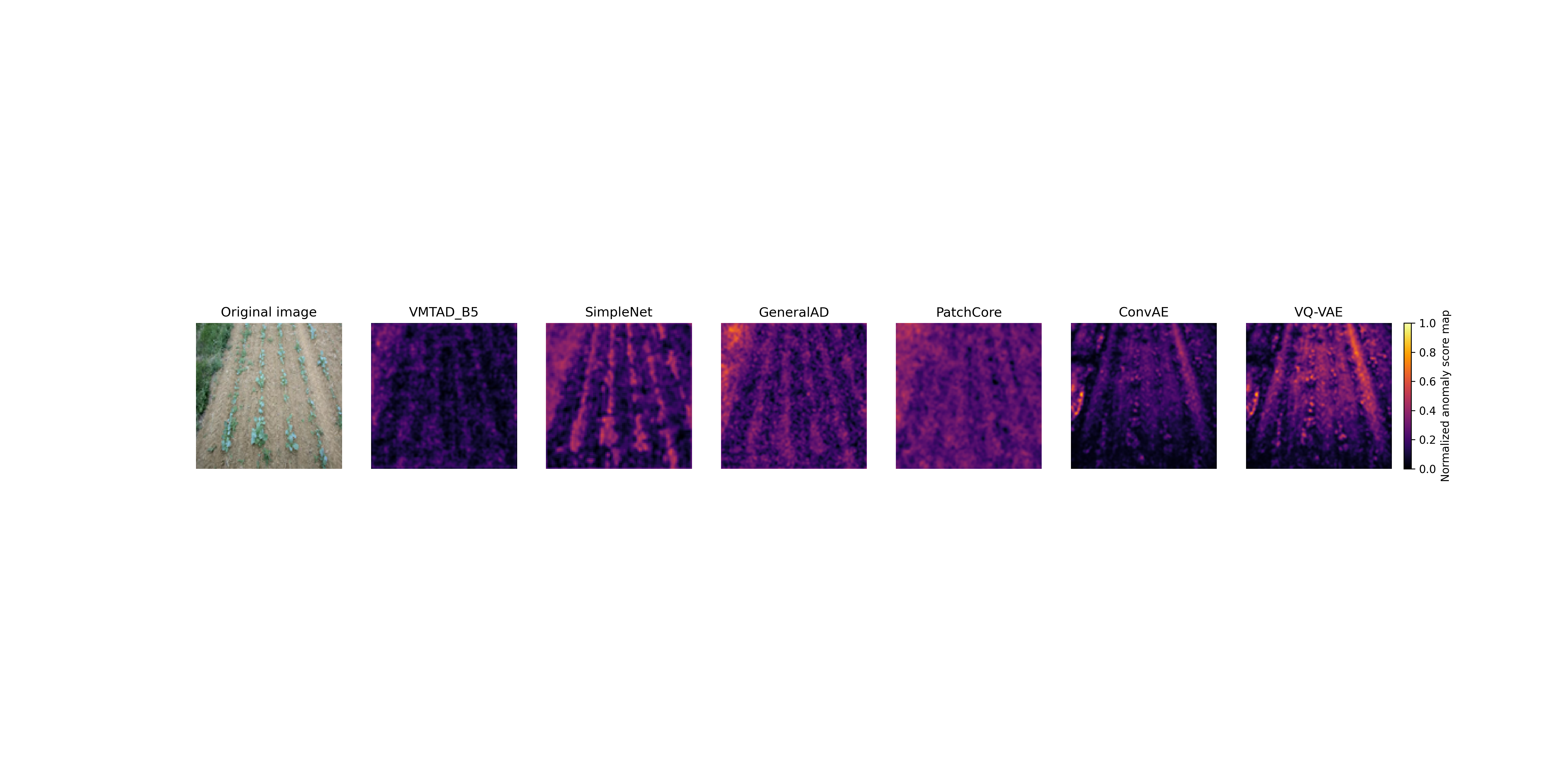}
		
		\caption{Comparison of anomaly score maps generated by different methods. The original image (left) contains multiple obstacles such as a irrigation pipe, wood log, human feet, fence. The last image does not contain any obstacles. The irrigation pipe was captured by a static rover, while the other images were taken by a moving rover. The inferno color map was used for its perceptual linearity.}
		\label{fig: score method}
	\end{figure*}

	Figure \ref{fig: score method} illustrates a visual comparison of the generated anomaly maps. High-intensity areas were observed at the location of obstacles (irrigation pipes, wood, human presence) in the VMTAD outputs. While CAE and VQ-VAE methods generated high anomaly scores for pipes, human feet or wooden logs were not highlighted. Lower anomaly scores were recorded by VMTAD on normal elements such as soil and plants compared to the SimpleNet, PatchCore or GeneralAD methods.
	
\begin{figure*}[!ht]
		\centering
		% Answer: [trim={left bottom right top},clip]
		\includegraphics[trim={0cm 0 2cm 0cm},clip,width=0.99\linewidth]{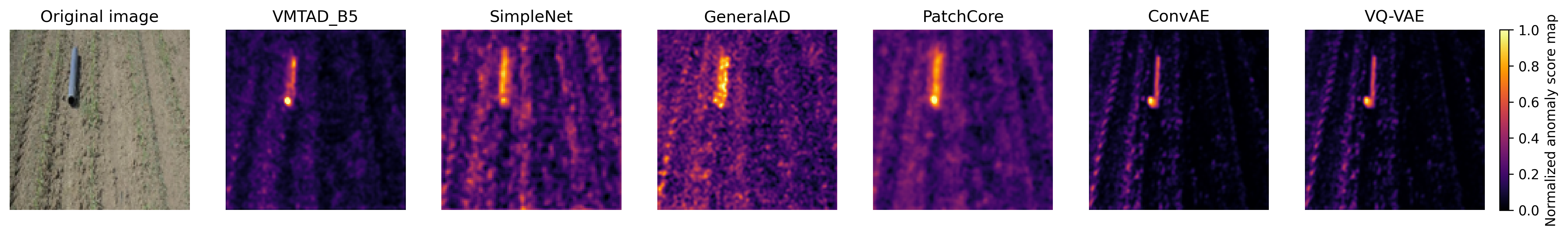}
		\includegraphics[trim={0cm 0 2cm 1cm},clip,width=0.99\linewidth]{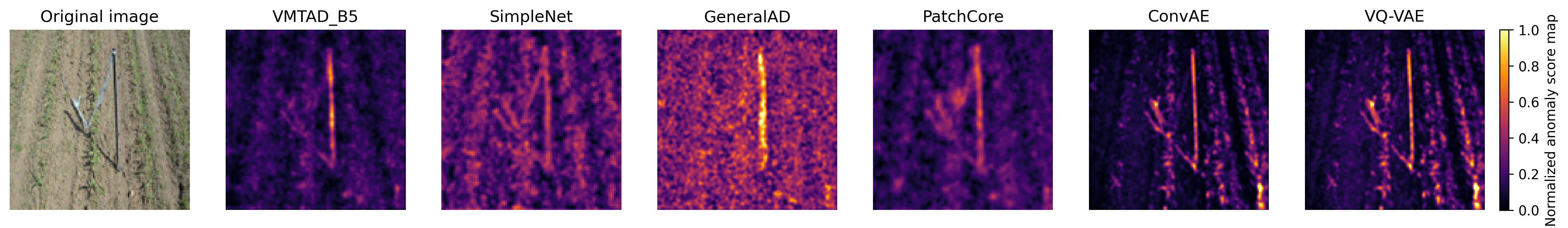}
		\includegraphics[trim={0cm 0 2cm 1cm},clip,width=0.99\linewidth]{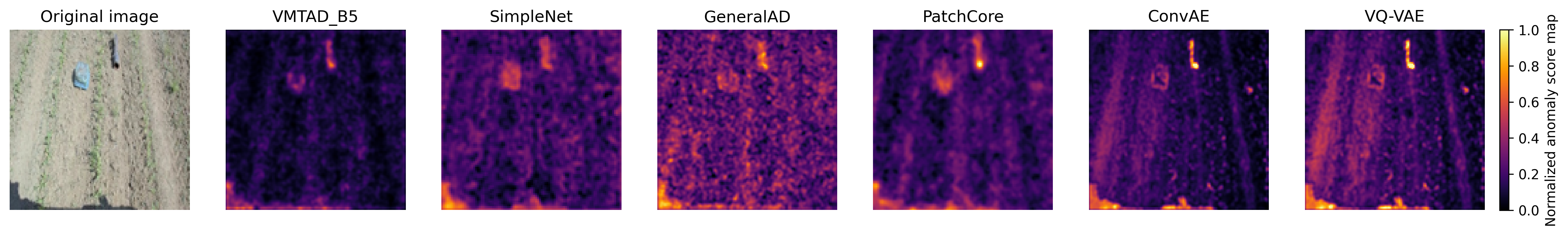}
		\includegraphics[trim={0cm 0 2cm 1cm},clip,width=0.99\linewidth]{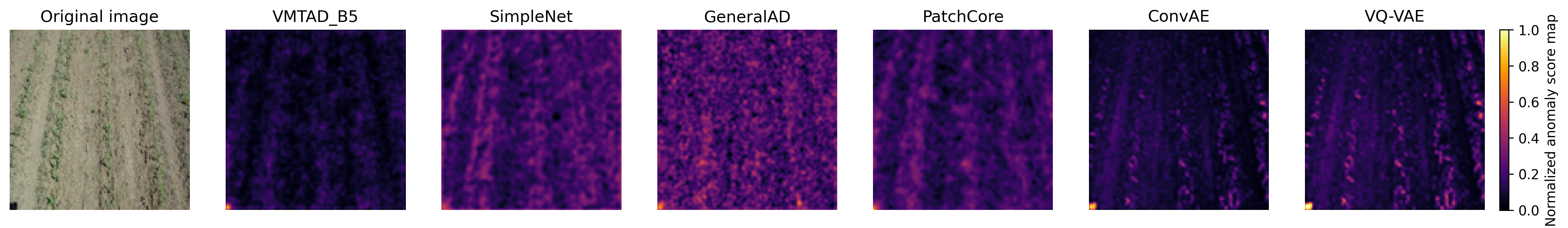}
		
		\caption{Comparison of anomaly score maps generated by different methods on cross-dataset configuration. The original image (left) contains multiple obstacles such as pipe, fence, 	random waste. The last image does not contain any obstacles. All images  was captured by a moving  rover. The inferno color map was used for its perceptual linearity.}
		\label{fig: score cross method}
	\end{figure*}

	Similarly, Figure~\ref{fig: score cross method} provides a visual comparison of the anomaly maps generated during cross-domain evaluation (trained on rapeseed and tested on corn). While VMTAD demonstrated strong overall performance, row 4 highlights a specific limitation: small shadows can trigger isolated, excessively high anomaly values. Interestingly, larger shadowed areas do not pose a difficulty for VMTAD, whereas they lead to significant performance degradation for SimpleNet, GeneralAD, ConvAE, and VQ-VAE.
	
	\subsection{Supplementary Study of the Memory Module}
	The impact of the FIFO queue size ($N$) and the memory decoder length ($K$) was quantified \ref{tab: FIFO} and \ref{tab: length decoder}.
	\begin{itemize}
		\item \textbf{FIFO Queue Size:} The addition of memory elements in FIFO queue increased the detection AUROC from 0.970 ($N=1$) to a maximum of 0.973 ($N=2$). Inference time increased progressively with $N$, rising from 61 ms for $N=1$ to 68 ms for $N=8$. The segmentation AUROC remains stable regardless of the number of memory elements. N=0 represents VTMAD without a memory module. 
		
		\item \textbf{Decoder Length:} Detection performance reached a peak of 0.973 for $K=2$ before declining for higher values. Segmentation remained stable at 0.997 for $K \in [1, 6]$.
	\end{itemize}
	
	\begin{table}[ht]
		\caption{AUROC performance as a function of FIFO queue size. In \textbf{bold} the best result and \underline{underline} the second-best result.}
		\label{tab: FIFO}
		\centering
		\begin{tabular}{cccc}
			\hline
			\(N\)        &  Det. $\uparrow$     & Seg. $\uparrow$   & Time (ms) $\downarrow$ \\ \hline
			\(\emptyset\) & 0.967               & \underline{0.996} & \textbf{52}   \\
			1            & \underline{0.970}    & \textbf{0.997}    & \underline{61}            \\
			2            & \textbf{0.973}       & \textbf{0.997}    & 62            \\
			4            & \textbf{0.973}       & \textbf{0.997}    & 64            \\
			6            & \underline{0.971}    & \textbf{0.997}    & 66            \\
			8            & 0.970                & \underline{0.996} & 68            \\
			\hline
		\end{tabular}
	\end{table}
	
	\begin{table}[ht]
		\caption{AUROC performance as a function of memory decoder length. In \textbf{bold} the best result and \underline{underline} the second-best result.}
		\label{tab: length decoder}
		\centering
		\begin{tabular}{cccc}
			\hline
			K             &  Det. $\uparrow$    & Seg. $\uparrow$           & Time (ms) $\downarrow$ \\ \hline
			1             & \underline{0.971}   & \textbf{0.997}                   & \textbf{58} \\
			2             & \textbf{0.973}      & \textbf{0.997}          & \underline{64} \\
			4             & \underline{0.972}      & \textbf{0.997}                    &  75 \\
			6             & 0.970               & \textbf{0.997}                      &  87 \\
			8             & 0.966               & \underline{0.996}            &  98 \\
			\hline
		\end{tabular}
	\end{table}
	
	\subsection{Braking Distance and Latency Impact}
\begin{figure*}[!ht]
		\centering
		\includegraphics[width=1\linewidth]{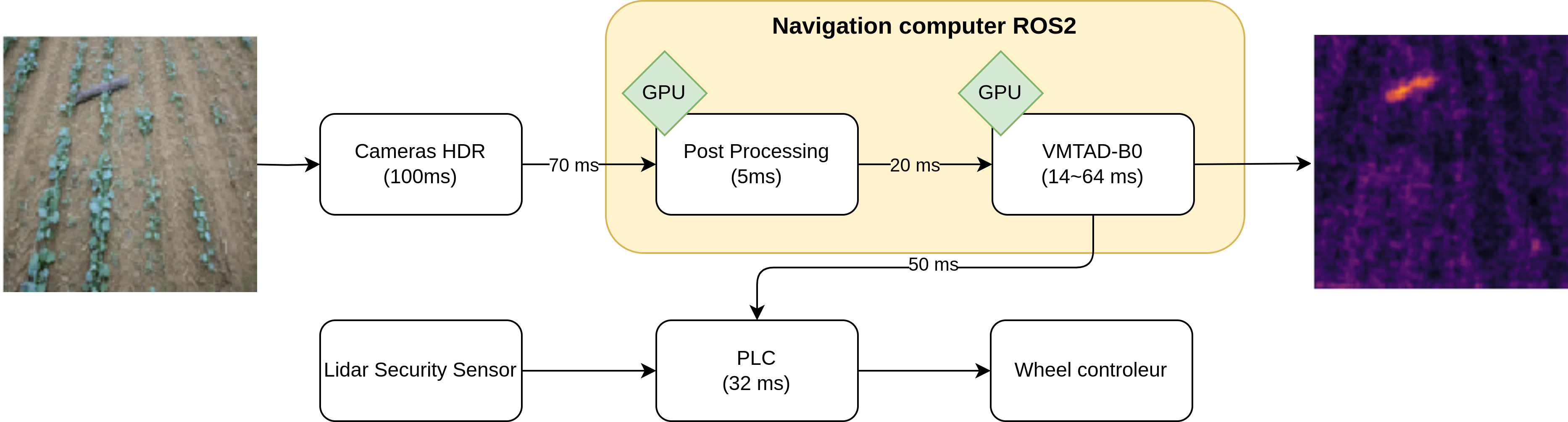}
		\caption{Grillon architecture. The inference time of each node is in () in each block. The cycle time of ROS2 is 20 ms, indicate by the arrow. The UDP cycle time is 50 ms.}
		\label{fig: grillon archi}
	\end{figure*}

	The analysis of the visual navigation loop Figure~\ref{fig: grillon archi} revealed a total reaction time ranging between 291 ms (B0 model) and 341 ms (B5 model). The total stopping distances calculated for the Grillon robot are detailed in Table \ref{tab:stopping_distance}:
	\begin{itemize}
		\item At a speed of 0.5 m s$^{-1}$, the stopping distance was 0.20 m with VMTAD-B0 and 0.22 m with VMTAD-B5.
		\item At the maximum speed of 1.2 m s$^{-1}$, the distance reached 0.57 m (B0) and 0.63 m (B5).
	\end{itemize}
	At this speed of 1.2 m s$^{-1}$, the vehicle traveled a distance of up to 0.41 m before the braking command was transmitted to the wheels.
	
	\begin{table}[ht]
		\centering
		\caption{Total stopping distance of the Grillon rover as a function of speed and the inference model used.}
		\label{tab:stopping_distance}
		\begin{tabular}{l c c} 
			\toprule
			\textbf{Model} & \textbf{Speed} & \textbf{Total Stopping} \\ 
			& \textbf{($m.s^{-1}$)} & \textbf{Distance (m)} \\ 
			\midrule
			\multirow{3}{*}{\begin{tabular}{@{}l@{}}\textbf{VMTAD-B0}\\(Latency: 291 ms)\end{tabular}} 
			& 0.5 & {0.20} \\ 
			& 1.0 & {0.44} \\ 
			& 1.2 & {0.57} \\ 
			\midrule
			\multirow{3}{*}{\begin{tabular}{@{}l@{}}\textbf{VMTAD-B5}\\(Latency: 341 ms)\end{tabular}} 
			& 0.5 & {0.22} \\ 
			& 1.0 & {0.49} \\ 
			& 1.2 & {0.63} \\
			\bottomrule
		\end{tabular}
	\end{table}

	\section{Discussion}
	\label{sec:discussion}
	
	\subsection{Comparative Analysis of Model Performance}
	As shown in Table~\ref{tab:results_rapseed}, the VMTAD-B5 variant demonstrates a substantial improvement over the reconstruction-based approaches commonly used in agriculture. While CAE, VQ-VAE, and MemAE \cite{s22103608} exhibit extremely low latency (4--5~ms), their detection fidelity may fall short for high-precision safety tasks. Within this group, VQ-VAE achieved the most balanced results (0.775 Det. AUROC), whereas CAE and MemAE struggled to exceed the 0.77 AUROC. This performance gap can be attributed to two primary factors. First, the identical shortcut problem \cite{ADTR_you} frequently affects convolutional models, even those equipped with memory, by allowing them to reconstruct anomalies too accurately. Transformers mitigate this issue, ensuring higher reconstruction errors for anomalous regions. Second, the attention mechanism provides Transformers with a global image context, whereas CNNs remain limited by a local receptive field. Conversely, CNN-based models offer a distinct advantage in terms of computational efficiency due to the parallelizable nature of convolutions and their linear spatial complexity, $O(H\times W)$.
	
	Beyond agricultural methods, VMTAD models offer competitive performance compared to industrial SOTA frameworks. While PatchCore \cite{Roth_2022_CVPR} achieves the highest detection score (0.977), VMTAD-B5 follows closely (0.973) and outperforms GeneralAD (0.960). This indicates that VMTAD is highly comparable to methods utilizing large-scale feature memory banks. Regarding anomaly segmentation, VMTAD-B5 achieves a high-level AUROC of 0.997, which is critical for agricultural robotics to facilitate the prediction of avoidance trajectories. This result highlights an effective trade-off: although not consistently superior in detection, the integration of FIFO memory within the dynamic context of autonomous weeding grants the model essential temporal awareness. This temporal propagation, combined with a global understanding, provides a specialized advantage over models like GeneralAD \cite{GeneralAD_cite}, which rely solely on static context, or PatchCore, which depends on extensive feature bank comparisons.
	
	\subsection{Memory impact}
	The results show Table~\ref{tab: FIFO} indicate that adding memory improves detection performance from 0.967 (no memory) to a peak of 0.973 at \(N=2\), while segmentation remains consistently high between 0.996 and 0.997. However, increasing \(N\) also increases the inference time from 52 ms without memory to 68 ms with \(N=8\). Thus, a FIFO queue size of 2 appears to provide the optimal balance between performance gains and computational efficiency. We also show that adding elements to the FIFO queue has a linear impact on computation time, as demonstrated by section \ref{Method}. 
	
	The results presented in Table~\ref{tab:length decoder} indicate that segmentation performance improves slightly with longer memory decoders (peaking at 0.997 for \(K\in[1,2,3,4]\), while detection performance is optimal for \(K=2\) (0.973). Since the inference time increases linearly with \(K\), a memory decoder length of 2 offers the best tradeoff between robust detection, high segmentation, and moderate computational cost.

	\subsection{Computational Efficiency and Real-Time Deployment}
	The performance of industrial methods often comes at a high computational cost. GeneralAD exhibits a latency of 101~ms, which exceeds requirements for real-time field deployment. In contrast, VMTAD-B5 provides superior segmentation accuracy at 64~ms. VMTAD-B0 variant provides a highly optimized balance, delivering near-perfect segmentation ($0.996$) with a latency of only 14~ms, making it more suitable for high-speed robotic inspection. PatchCore and SimpleNet \cite{Liu_2023_CVPR} are very fast due to their bank size optimized for PatchCore at 0.1\% of the dataset and the high parallelization of SimpleNet.
	
	\subsection{Cross-Domain Generalization Challenges}
	Evaluation on the corn dataset highlights the challenges of domain shift in precision agriculture. While SimpleNet, GeneralAD, and PatchCore perform well in detection, VMTAD variants show a decrease in detection AUROC. Nevertheless, the segmentation scores remain high, suggesting that obstacles are still correctly localized. 
	
	This performance gap can be attributed to the fact that SimpleNet and GeneralAD utilize self-supervised representation learning to recognize abstract deviations from normality. These methods are inherently more robust to noise and domain shifts. In contrast, VMTAD relies on reconstructing the input, this imposes a narrower definition of normality, as the model must reconstruct pixel-level details rather than just identifying feature-level patterns. Consequently, even minor deviations from the training distribution result in significantly higher reconstruction errors, making VMTAD more sensitive to environmental variance compared to the broader, more generalized boundaries learned by SimpleNet. However, this improved generalization comes at the cost of good detection accuracy in training domains, as the results show. 
	
	This analysis is confirmed by the qualitative analysis shown in Figure \ref{fig: score cross method}. VMTAD performs well in anomaly detection and object segmentation. However, small shadow areas, such as the one in the lower left of row 4, are identified as anomalies. In contrast, the larger shadow area in row 3 is not identified as an anomaly. We attribute this to VMTAD’s more restrictive definition of normality and the absence of such cases in the rapeseed dataset. It is also worth noting that industrial models are less sensitive to this situation, yet more sensitive to large shadow areas. 
	
	\subsection{Qualitative Localization and Environment Modeling}
	Visual analysis of anomaly score maps in Figure. \ref{fig: score method} demonstrates VMTAD's superior ability to model complex semantic textures. Agricultural baselines often fail to detect challenging objects like humans or logs, whereas VMTAD maintains low scores for non-abnormal areas such as soil and plants. Industrial models like SimpleNet or GeneralAD frequently misclassify soil with high anomaly values. Only PatchCore offers a similar quality, though it still produces higher anomaly scores for soil than VMTAD. We can again associate the low noise score with the fact that VMTAD has a temporal and global understanding, which is not the case with other models, allowing it to better reconstruct the environment.
	
	\subsection{System Latency and Safety Implications}
	The braking performance of the Grillon rover is influenced by the total latency of the perception pipeline. While VMTAD-B0 reduces the stopping distance compared to VMTAD-B5, the model's inference time is only one component of the reaction time. The system architecture introduces significant fixed latencies—100~ms for camera acquisition and 70~ms for data transfer—which remain constant regardless of the model speed. These architectural bottlenecks contribute more to the total delay (291--341~ms) than the choice of model.
	
	At a speed of 1.2~m s$^{-1}$, the rover travels up to 0.41~m before a braking command is transmitted in the worst-case scenario. Therefore, while optimizing inference time is important, substantial gains in safety require addressing the entire hardware and software architecture. A trade-off is required: VMTAD-B0 is preferable for high-speed operations to minimize risk, whereas VMTAD-B5 is suitable for low-speed tasks where superior detection accuracy is prioritized.

	\section{Conclusion}
	
	This study introduced VMTAD, a method that advances the state-of-the-art in agricultural anomaly detection by harmonizing Transformer architectures with temporal memory. The scientific rationale behind this approach addresses a fundamental gap in robotic vision: the inability of static reconstruction models to account for the temporal dependencies inherent in continuous motion. By capturing these dynamics, the framework transforms anomaly detection from a frame-by-frame analysis into a coherent spatio-temporal process.
	
	A critical finding of this study is the indispensable role of high-fidelity detection AUROC in autonomous pipelines. While traditional reconstruction-based methods (CAE, VQ-VAE, MemAE) offer rapid inference, their lower detection scores introduce unacceptable risks of false negatives, which in a field setting could lead to untreated crop diseases or missed structural failures. Our VMTAD-B5 variant bridges this gap, achieving a detection AUROC of 0.973 (comparable to heavy industrial models) while maintaining the operational speed necessary for real-time robotic response. This high detection reliability ensures that the system can accurately detecte obstacles without the prohibitive 101~ms latency of existing industrial SOTA models like GeneralAD or low segmentation score than PatchCore.
	
	Despite these advancements, the study reveals that the model remains sensitive to cross-domain generalization. The observed performance degradation when transitioning from rapeseed to corn highlights the persistent challenge of domain shift in precision agriculture, where biological variance is high. 
	
	Future research will focus on: (1) enhancing detection robustness against environmental noise and shift; (2) implementing multimodal data fusion to leverage depth or multispectral sensors for more resilient anomaly identification. Ultimately, securing high-precision detection at low latency is the cornerstone of reliable autonomous agriculture, contributing to more efficient and sustainable farming practices.

	\section*{Funding sources}
This work was supported by ANRT as part of a CIFRE (Industrial Research Training Contracts) doctoral research grant and by CYCLAIR.

	\bibliography{biblio}
	
\end{document}